\documentclass[nonacm,acmsmall]{acmart}

\AtBeginDocument{%
  }

\setcopyright{acmcopyright}
\copyrightyear{2018}
\acmYear{2018}
\acmDOI{XXXXXXX.XXXXXXX}
\acmConference[Conference acronym 'XX]{Make sure to enter the correct
  conference title from your rights confirmation email}{June 03--05,
  2018}{Woodstock, NY}
\acmPrice{15.00}
\acmISBN{978-1-4503-XXXX-X/18/06}

\usepackage{xspace}
\usepackage{xcolor}
\usepackage{bm}
\usepackage{wasysym}

\newcommand{\newtextsc}[1]{\textsc{#1}}
\newcommand{\sysname}{CORE\xspace}
\newcommand{\sorald}{\textsf{Sorald}\xspace}

\newcommand{\datasetpy}{\newtextsc{CQPy}}
\newcommand{\datasetpyhuman}{\newtextsc{CQPyUS}}
\newcommand{\datasetjava}{\newtextsc{SQJava}}

\newcommand\doublecheck{\textcolor{teal}{\checked\kern-0.5em\checked}}

\newcommand{\codeql}{CodeQL\xspace}
\newcommand{\sq}{\newtextsc{SonarQube}\xspace}
\newcommand{\turbo}{\newtextsc{GPT-3.5-Turbo}\xspace}

\newcommand{\gptfour}{\newtextsc{GPT-4}\xspace}

\newcommand{\code}[1]{\texttt{#1}}
\newcommand{\mytexttt}[1]{{\color{black}\texttt{#1}}}
\newcommand{\mytextttm}[1]{{\color{black}\texttt{#1}}}
\newcommand{\eq}{\texttt{\_\_eq\_\_}\xspace}
\newcommand{\hash}{\texttt{\_\_hash\_\_}\xspace}

\newcommand{\othersysname}[1]{\newtextsc{#1}\xspace}
\newcommand{\detete}[1]{{}}

\newcommand{\delete}[1]{{}}

\newcommand{\eqnotoverridden}{\newtextsc{Eq-Not-Overridden}\xspace}

\newcommand{\genprompt}{\newtextsc{Proposer Prompt}\xspace}

\newcommand{\checkerprompt}
{\newtextsc{Ranker Prompt}\xspace}

\newcounter{inlineenum}
\renewcommand{\theinlineenum}{\arabic{inlineenum}}
\newenvironment{inlineenum}
  {\unskip\ignorespaces\setcounter{inlineenum}{0}%
   \renewcommand{\item}{\refstepcounter{inlineenum}{{(\theinlineenum})~}}}
  {\ignorespacesafterend}

\newcommand{\prompt}[2]{%
    \begin{tcolorbox}[
        size=small,
        space to upper,
        colframe=black!80!white,
        colback=white,
        halign=left,
        valign=top,
        halign lower=flush left,
        title=#1
    ]
    #2
    \end{tcolorbox}%
}

\usepackage{color}
\definecolor{deepblue}{rgb}{0,0,0.5}
\definecolor{deepred}{rgb}{0.6,0,0}
\definecolor{deepgreen}{rgb}{0,0.5,0}

\definecolor{commentgreen}{RGB}{2,112,10}
\definecolor{eminence}{RGB}{108,48,130}
\definecolor{brickred}{RGB}{170,74,68}
\definecolor{teal}{RGB}{0,128,128}

\usepackage{listings}

\newcommand\pythonstyle{\lstset{
  language=Python,
  basicstyle=\footnotesize\ttfamily,
  columns=fullflexible,
  commentstyle=\color{commentgreen},
  keywordstyle=\color{blue},
  emph={int,char,double,float,unsigned,void,bool, class},
  emphstyle={\color{blue}},
  frame=single,
  breaklines=true,
  postbreak=\mbox{\textcolor{red}{$\hookrightarrow$}\space},
  showstringspaces=false,
  commentstyle=\color{gray}\upshape,
}}

\lstnewenvironment{python}[1][]
{
\pythonstyle
\lstset{#1}
}
{}

\newcommand\diffstyle{\lstset{
  language=Python,
  basicstyle=\footnotesize\ttfamily,
  columns=fullflexible,
  keywordstyle=\color{blue},
  emph={int,char,double,float,unsigned,void,bool, class},
  emphstyle={\color{blue}},
  frame=single,
  breaklines=true,
  postbreak=\mbox{\textcolor{red}{$\hookrightarrow$}\space},
  showstringspaces=false,
  keepspaces=true,
  commentstyle=\color{gray}\upshape,
}}

\lstnewenvironment{diff}[1][]
{
\diffstyle
\lstset{#1}
}
{}

\newcommand*\circled[1]{\tikz[baseline=(char.base)]{
            \node[shape=circle,fill,inner sep=0.5pt] (char) {\textcolor{white}{#1}};}}

\newcommand*\stage[1]{\tikz[baseline=(char.base)]{
            \node[shape=circle,fill,inner sep=0.5pt,color=brickred] (char) {\textcolor{white}{#1}};}}

\newcommand{\myparagraph}[1]{\par\vspace{2mm}\noindent\textbf{#1}}

\usepackage{amsmath,amsfonts,bm,mathrsfs}
\usepackage{algorithmic}
\usepackage{graphicx}
\usepackage{textcomp}
\usepackage{xcolor}
\usepackage{tikz}
\usepackage{booktabs}
\usepackage{url}
\usepackage{longtable}
\usepackage{framed}
\usepackage{multirow}
\usepackage{listings}
\usepackage[many]{tcolorbox}
\usepackage{mdframed}

\usepackage{graphbox}
\usepackage[export]{adjustbox}

\begin{document}

\title{Frustrated with Code Quality Issues? LLMs can Help!}

\author{Nalin Wadhwa}
\affiliation{%
  \institution{Microsoft Research}
  \country{India}
}
\author{Jui Pradhan}
\affiliation{%
  \institution{Microsoft Research}
  \country{India}
}
\author{Atharv Sonwane}
\affiliation{%
  \institution{Microsoft Research}
  \country{India}
}
\author{Surya Prakash Sahu}
\affiliation{%
  \institution{Microsoft Research}
  \country{India}
}
\author{Nagarajan Natarajan}
\affiliation{%
  \institution{Microsoft Research}
  \country{India}
}
\author{Atharv Sonwane}
\email{t-asonwane@microsoft.com}
\affiliation{%
  \institution{Microsoft Research}
  \country{India}
}

\author{Aditya Kanade}
\email{kanadeaditya@microsoft.com}
\affiliation{%
  \institution{Microsoft Research}
  \country{India}
}

\author{Suresh Parthasarathy}
\email{supartha@microsoft.com}
\affiliation{%
  \institution{Microsoft Research}
  \country{India}
}

\author{Sriram Rajamani}
\email{sriram@microsoft.com}
\affiliation{%
  \institution{Microsoft Research}
  \country{India}
}
 \email{larst@affiliation.org}

\begin{abstract}
As software projects progress, quality of code assumes paramount importance as it affects reliability, maintainability and security of software. For this reason, static analysis tools are used in developer workflows to flag code quality issues. However, developers need to spend extra efforts to revise their code to improve code quality based on the tool findings. In this work, we investigate the use of (instruction-following) large language models (LLMs) to assist developers in revising code to resolve code quality issues.

We present a tool, \sysname (short for COde REvisions), architected using a pair of LLMs organized as a duo comprised of a proposer and a ranker. Providers of static analysis tools recommend ways to mitigate the tool warnings and developers follow them to revise their code. The \emph{proposer LLM} of \sysname takes the same set of recommendations and applies them to generate candidate code revisions. The candidates which pass the static quality checks are retained. However, the LLM may introduce subtle, unintended functionality changes which may go un-detected by the static analysis. The \emph{ranker LLM} evaluates the changes made by the proposer using a rubric that closely follows the acceptance criteria that a developer would enforce. \sysname uses the scores assigned by the ranker LLM to rank the candidate revisions before presenting them to the developer.

We conduct a variety of experiments on two public benchmarks to show the ability of \sysname:
\circled{1}~to generate code revisions acceptable to both static analysis tools and human reviewers (the latter evaluated with user study on a subset of the Python benchmark),
\circled{2}~to reduce human review efforts by detecting and eliminating revisions with unintended changes,
\circled{3}~to readily work across multiple languages (Python and Java), static analysis tools (CodeQL and SonarQube) and quality checks (52 and 10 checks, respectively),
and
\circled{4} to achieve fix rate comparable to a rule-based automated program repair tool but with much smaller engineering efforts (on the Java benchmark).
\sysname could revise 59.2\% Python files (across 52 quality checks) so that they pass scrutiny by both a tool and a human reviewer. The ranker LLM reduced false positives by 25.8\% in these cases. \sysname produced revisions that passed the static analysis tool in 76.8\% Java files (across 10 quality checks) comparable to 78.3\% of a specialized program repair tool, with significantly much less engineering efforts.
\end{abstract}



\maketitle

\section{Introduction}
As software projects progress, assessing reliability, maintainability and security of software assumes paramount importance. Quality of code plays a big role in ensuring these  objectives~\cite{duvall2007continuous, fowler2018refactoring}. Many static analysis tools like CodeQL~\cite{DBLP:conf/ecoop/AvgustinovMJS16}, Coverity~\cite{coverity}, FindBugs~\cite{findbugs}, PMD~\cite{pmd} and SonarQube~\cite{sonarsa} are used in developer workflows to flag code quality issues. However, developers need to spend extra efforts to revise their code to improve code quality based on the tool findings~\cite{7962383,vassallo2020developers}.

Recognizing the value of static analysis tools in improving code quality, many approaches~\cite{FootPatch,senx,costea2023hippodrome,sonarcube,Liu:mining,8667970,bader2019getafix,bavishi2019phoenix,someoliayi2022sorald,jain2023staticfixer} use them to detect and localize violations of static checks. To fix the violations, they use either manually designed symbolic program transformations~\cite{FootPatch,senx,someoliayi2022sorald,costea2023hippodrome}, mine symbolic patterns from commit data~\cite{sonarcube,Liu:mining,8667970,bader2019getafix,bavishi2019phoenix} or learn them from synthetically generated data~\cite{jain2023staticfixer}. The code or fix generation capabilities of these symbolic approaches are limited by the space of supported patterns. Learning-based approaches~\cite{tufano2019empirical,zhu2021syntax,jiang2021cure,ye2022neural} try to overcome this limitation by training neural models to map buggy programs to their fixed versions. However, similar to pattern-mining approaches, these require bug-fixing data for training and it limits the types of bugs they can fix.
Setting up these systems and supporting a different programming language, a new quality check or another static analysis tool incurs significant engineering costs. These factors prevent the wide-spread adoption of these automated program repair (APR) tools. 

\begin{figure}[t]
\includegraphics[scale=0.215]{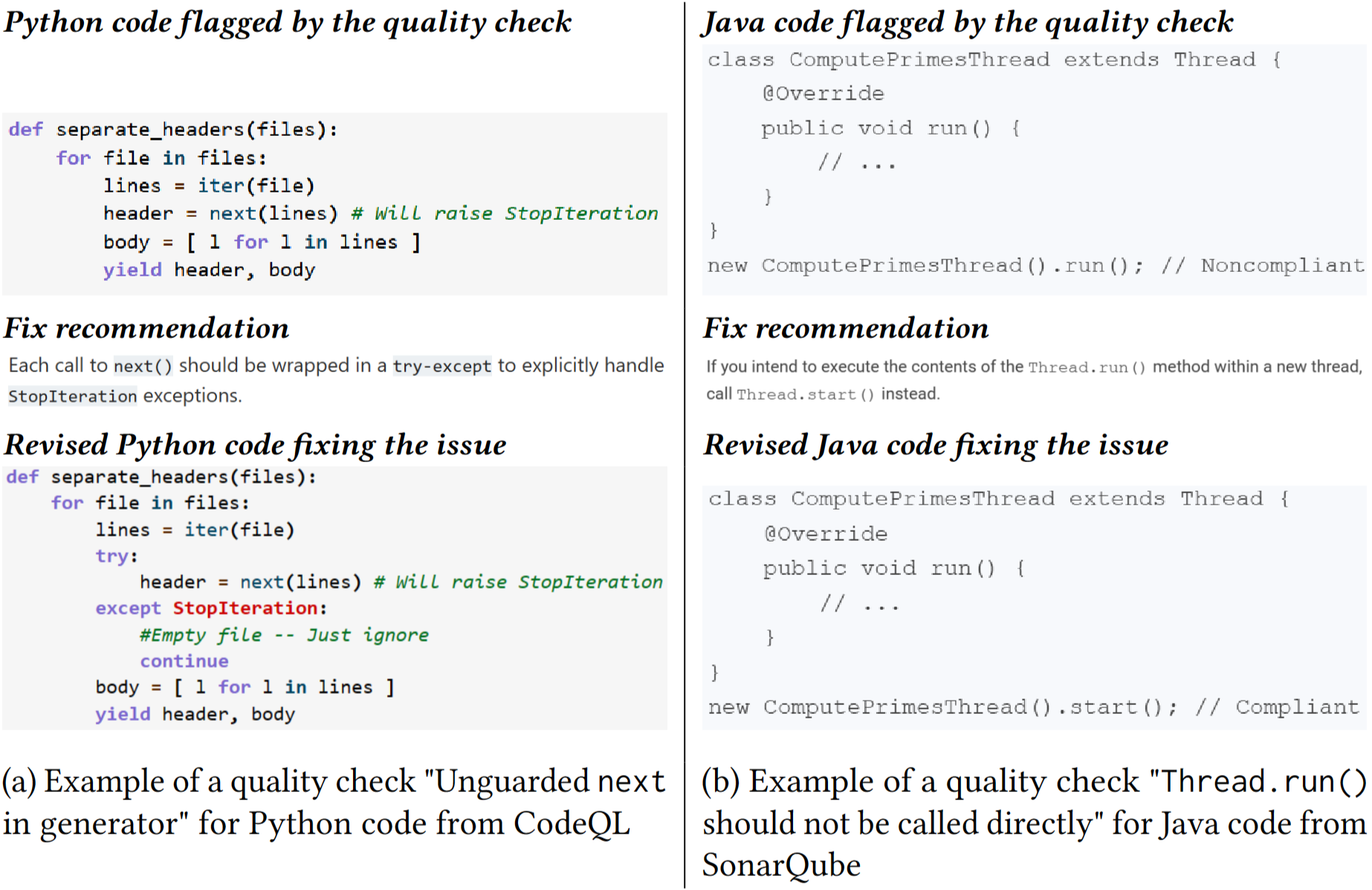}
\caption{Examples of quality checks, fix recommendations, and code before and after following the fix recommendations for CodeQL and SonarQube tools for Python and Java languages respectively.}
\label{fig:checks-and-fixes}
\end{figure}

Providers of static analysis tools recommend ways to mitigate the tool warnings. Developers can follow them to manually revise their code when warnings are raised. Figure~\ref{fig:checks-and-fixes} illustrates two quality checks (a) and (b) from two tools: CodeQL applied to Python code\footnote{\url{https://codeql.github.com/codeql-query-help/python/py-unguarded-next-in-generator/}}  and SonarQube applied to Java code\footnote{\url{https://rules.sonarsource.com/java/RSPEC-1217/}}. At the top are code snippets with quality issues. The natural-language fix recommendations for the quality checks are shown in the middle and the revised code that can be obtained after manually following the fix recommendations are shown at the bottom.

The APR tools try to learn mapping between the original and revised snapshots of the code. \emph{To avoid the limitations of APR tools outlined above, we propose to instead make direct use of the clear and concise natural-language instruction (fix recommendation) supplied by the tool providers.} The emergence of large language models (LLMs) (e.g.,~\cite{brown2020language,chen2021evaluating,chowdhery2022palm,touvron2023llama,openai2023gpt4}) offers an opportunity to make this possible. LLMs are large neural networks that capture generative distributions of natural languages and source code. These models are trained on very large data in unsupervised manner. Instruction-tuning~\cite{ouyang2022training} enhances their utility by finetuning the base LLMs to comprehend and follow natural language instructions. As we show in this paper, it is possible to \emph{instruct} state-of-the-art LLMs to revise a piece of code directly using natural language instructions. These models can sample a variety of code conditioned on instructions, that too without any additional training or finetuning. Unlike symbolic program transformations and neural models trained on specific datasets, they are not limited by the space of patterns or bug-fixing data used for training. This eliminates the need to expend the efforts required in designing symbolic transformation systems or training specialized neural models, and saves on engineering efforts. 

We try to realize the promise of LLMs to resolve code quality issues flagged by static analyses in a tool called \sysname (short for COde REvision). \sysname is architected using a pair of LLMs organized as a duo comprised of a proposer and a ranker. The \emph{proposer LLM} of \sysname takes a fix recommendation and applies it to a given source-code file to generate candidate code revisions. The candidates which pass the static quality checks are retained. However, the LLM may introduce subtle, unintended functionality changes which may go un-detected by the static analysis. The \emph{ranker LLM} evaluates the changes made by the proposer using a rubric that closely follows the acceptance criteria that a developer would enforce. \sysname uses the scores assigned by the ranker LLM to rank the candidate revisions before presenting them to the developer.

We evaluate \sysname on two public benchmarks: CodeQueries~\cite{sahu2022learning} and \sorald~\cite{someoliayi2022sorald}. CodeQueries is a benchmark of Python files with quality issues flagged by one of the 52 common static checks applied by the CodeQL tool. \sorald comprises of Java repositories with quality issues flagged by one of the 10 common static checks applied by the SonarQube tool. Both the datasets contain code from public GitHub repositories and are representative of real-world quality issues. 

We conduct a variety of experiments on these benchmarks to show the ability of \sysname:
\circled{1} to generate code revisions acceptable to both static analysis tools and human reviewers (the latter evaluated with user study on a subset of the Python benchmark),
\circled{2} to reduce human review efforts by detecting and eliminating revisions with unintended changes,
\circled{3} to readily work across multiple languages (Python and Java), static analysis tools (CodeQL and SonarQube) and quality checks (52 and 10 checks, respectively),
and
\circled{4} to achieve fix rate comparable to a rule-based automated program repair tool, Solard, but with much smaller engineering efforts (on the Java benchmark).

We obtain promising results that bear witness to practical utility of \sysname using \turbo~\cite{ouyang2022training} as the proposer LLM and \gptfour~\cite{openai2023gpt4} as the ranker LLM.
\sysname could revise 59.2\% Python files (across 52 quality checks) so that they pass scrutiny by both a tool and a human reviewer. The ranker LLM reduced the false positive rate by 25.8\% in these cases. \sysname produced revisions that passed the static analysis tool in 76.8\% Java files (across 10 quality checks) compared to 78.3\% of the specialized program repair tool Solard~\cite{someoliayi2022sorald}, but with significantly much lesser engineering efforts. The authors of Solard state that ``The design and implementation of SORALD already represents 2+ years of full time work.''~\cite{someoliayi2022sorald}, whereas we were able to apply \sysname on the Solard benchmark with a few days of efforts by a couple of authors. 

There is growing interest in using LLMs in program repair. Many existing techniques~\cite{fan2022automated, liventsev2023fully,xia2023automated,xia2023conversational} aim at fixing bugs characterized by failing test cases. In comparison, we focus on fixing quality issues that are discovered statically and do not have accompanying unit tests for validation. The techniques that repair statically detected errors~\cite{joshi2022repair,jin2023inferfix,pearce2022examining} either target syntactic or simple semantic errors~\cite{joshi2022repair}, finetune an LLM on specially designed prompts~\cite{jin2023inferfix} or use less powerful models and prompting~\cite{pearce2022examining}. We target a wide range of code quality issues using an instructed-tuned LLM which supports powerful prompting without any finetuning.

We make the following contributions in this paper:
\begin{itemize}
\item We identify an emerging opportunity of using instruction-following LLMs to assist developers in resolving code quality issues.
\item We present a system, \sysname, to evaluate this opportunity. We design a multi-step protocol wherein one LLM proposes the code revisions, which are filtered by applying the static analysis and further ranked using a ranker LLM, before they are presented to the developer.
\item We conduct extensive experimentation to evaluate the acceptability of the revisions produced by \sysname, its ability to control false positives, generalizability to different languages, tools and checks, and its performance compared to a specialized program repair tool. Our results show that \sysname is a promising step in bringing LLMs to the help of developers in resolving code quality issues. We also identify opportunities for further improvements. 
\item We will to release our code and data at \url{https://aka.ms/CORE_MSRI}.
\end{itemize}

\section{Overview}
\label{sec:overview}
\begin{figure*}[t!]
\centering
\includegraphics[scale=0.38]{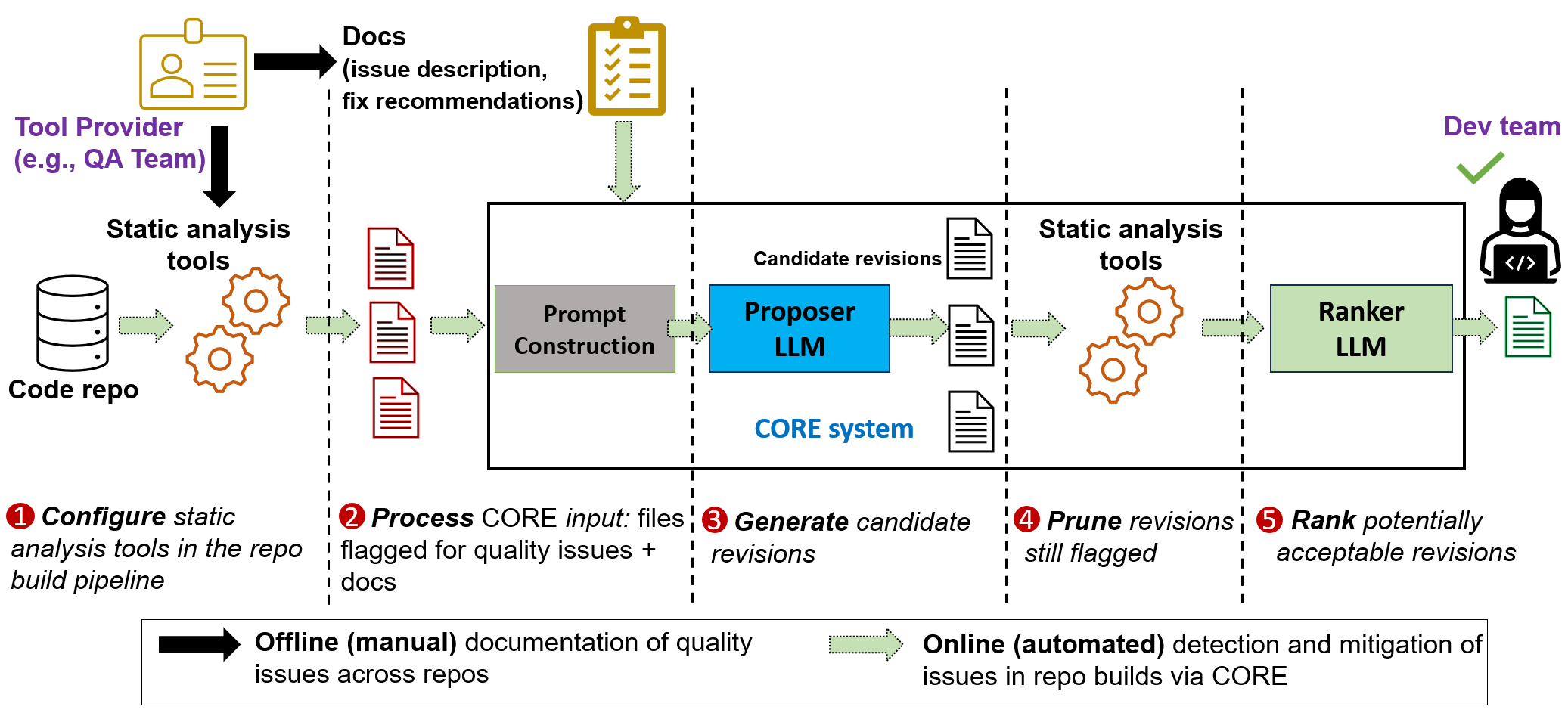}
\caption{\sysname pipeline: Code quality issues (static checks) across software repositories are documented by the tool provider. \sysname is integrated in the repo build pipeline that also runs the suite of static analysis checks. The flagged source files and the documentation are fed as input to the \sysname system to automatically produce source file revisions that address the quality issues. The candidate revisions that pass the static checks are further assessed and ranked by a ranker LLM to prevent surfacing spurious fixes to the developer.}
\label{fig:core}
\end{figure*}

Our goal is to (1) automate code quality revisions in software engineering workflows, which typically comprise large-scale code repositories and various code quality control checks; (2) by taking static analysis tools and documentation of quality checks (natural language instructions) as input; (3) with minimal developer intervention. This section gives an overview of the architecture of \sysname using an example code quality issue resolution scenario.

Consider the Python code snippet shown in Code 1 below. The~\mytexttt{PersistentDict} class derives from \mytexttt{dict} class, adding \mytextttm{\_filename} and \mytextttm{\_transact} attributes of its own. It does not override the \mytextttm{\_\_eq\_\_} method. This is an instance of poor code quality that can cause errors: when two objects of \mytexttt{PersistentDict} class are compared, the subclass attributes get ignored. This example code gets flagged by the \codeql\ tool with the \eqnotoverridden\ warning \cite{eqnotoverridden}. 

Software engineering workflows involve such quality checks for readability, maintainability, security, etc. Accompanying these checks are the   guidelines (natural language instructions) for fixing the issues in the source files that are flagged by static analysis tools. For the \eqnotoverridden\ check, the \codeql\ manual page on the web\footnote{\url{https://codeql.github.com/codeql-query-help/python/py-missing-equals/}} states \textit{``A class that defines attributes that are not present in its superclasses may need to override the \mytextttm{\_\_eq\_\_}() method (\mytextttm{\_\_ne\_\_}() should also be defined)''}.  Tool developers or quality assurance (QA), security and compliance teams in organizations write the static quality checks and documentation (fix recommendations), and the repository owners (dev team) are responsible for fixing the quality issues based on the provided guidelines. 

\begin{python}[caption={Example Python code with \eqnotoverridden issue.},captionpos=b]
class PersistentDict(dict):
    """A class that persists a dict to a file. This class behaves like a dict and adds new functionality to store the dict to a file when writing."""
    def __init__(self, filename, load=True):
        self._filename = os.path.abspath(filename)
        if load: self._load()
        self._transact = False

    @property
    def filename(self):
        'The filepath to write'
        return self._filename
\end{python}

In the following, we present an overview of the \sysname\ pipeline that is configured with the inputs from the tool provider along with the source-code files flagged by the tools, and produces automatic revisions of the source files to address the issues. 

\noindent
\stage{1} \textbf{Configuring the CORE pipeline:}
\sysname, shown in Figure~\ref{fig:core}, is a generic pipeline for code revisions. In order to configure \sysname to process
 the \eqnotoverridden{} issue, the tool provider supplies two types of information to \sysname:
 \begin{inlineenum}
    \item the static analysis tool (e.g., \codeql) and the check itself (e.g., a \texttt{.ql} file),
    \item a description of the code quality issue and instructions to fix the issue in natural language.
\end{inlineenum}
In our evaluation, we obtain the description of the quality issues and the instructions to fix them from online documentation.  Once \sysname\ is configured with the two types of information, it can automatically process static analysis reports corresponding to this issue, along with the files containing the code that needs to be revised to fix the issue, and propose candidate revisions to the code. 
\begin{framed}\noindent\textit{The aforementioned manual configuration for a code quality issue is a one-time, offline step, that serves to produce revisions (in an automated, online fashion) for the issue arising in various repositories, thereby automating the repetitive task of resolving the code quality issues.}\end{framed}
We describe the components of the \sysname\ pipeline next.

\noindent
\stage{2} \textbf{Constructing prompt:} 
This component takes in the static analysis report, the flagged source file, and the documentation for the issue (see above), and constructs a ``prompt'' for querying the large language model (LLM). A \emph{prompt} encodes the natural language instruction to solve a particular task and optionally additional information that the model might use to perform the task such as hints (e.g., lines of interest in the file), constraints (e.g., ``{do not modify parts of code unrelated to the issue}''), and demonstrations (e.g., an example code snippet with \eq not overridden and its revised version). Typically, each LLM has a limit on the prompt size (also referred as context size) it supports in terms of the number of tokens. In \sysname, the prompt encapsulates the description of the quality issue, fix suggestions, and the line(s) of code the static analysis report attributes the issue to. In addition, \sysname employs other relevant information that may be helpful to fix the issue, such as fetching relevant blocks of code derived from static analysis reports. The details of prompt construction are presented in Section \ref{sec:design}.\\
\noindent
\stage{3} \textbf{Generating candidate revisions using Proposer LLM:} 
The proposer LLM takes as input the constructed prompt in natural language along with the code flagged for the quality issue and outputs potential code revisions. We use \turbo, which is a state-of-the-art LLM for code generation, in our experiments. \turbo supports large prompt sizes (up to 4000 tokens). This lets us input the entire source-code file for many cases.
For very large files, we give the largest context block admissible (e.g., the entire method or class surrounding the lines of interest) by the prompt size (details in Section \ref{sec:design}). The output code (a block or a full file, as the case may be) is then patched back to the original file. We sample 10 candidate revisions for each input file.\\
\noindent
\stage{4} \textbf{Pruning revisions with the configured tool:} 
We run the static analysis check (that \sysname is configured with in \stage{1}) against the candidate revisions and filter out the ones where the code quality issue continues to persist (i.e., non-zero violations detected). In this process, we de-duplicate identical code revision suggestions as well as ensure the syntactic validity of code revisions, and reject the ones with syntax errors.\\
\noindent
\stage{5} \textbf{Ranking admissible candidates using Ranker LLM:} 
The static analysis tool could pass revisions that are not acceptable to developers, such as introducing unintended changes in the code (e.g., a revision that overrides the \eq method properly, but alters the implementation of \hash unnecessarily). The impressive editing and generative ability of state-of-the-art LLMs also means that they can make alterations (often subtle, but changing the semantics) to existing code even if they are explicitly instructed \textit{not} to do so. Furthermore, if the quality of the static check (usually some form of pattern matching) itself is poor, it is even more important to ensure incorrect revisions are not surfaced to the developer who will eventually accept or reject them. To reduce the burden on the developers, and to improve the acceptance rate of the surfaced fixes, we employ another LLM (\gptfour) to score the candidate revisions in the order of their likelihood of acceptance. To instruct the LLM to do so, we rely on the same rubric, i.e., instructions for issue resolution supplied in \stage{1}, to construct a prompt for the LLM. In particular, we give the candidate revision that passed the static analysis tool (diff with the original source code) along with the rubric in natural language, as input to the Ranker LLM and ask it to assign an ordinal score in a range (from \textit{strong accept} to \textit{strong reject}). We use this score to rank the potential candidates for a file. The details of the ranking strategy are discussed in Section \ref{sec:design}.  \\
~\\
\noindent
\textbf{Sample output:} 
Code snippet shown below is a sample output generated by \sysname  for Code 1. Due to lack of space, instead of showing the full code, we display the outputs in the standard diff format with respect to Code 1. 

\begin{mdframed}[linewidth=0pt]
\begin{diff}[caption={A correct revision of Code 1 generated by \sysname.},captionpos=b]
@@ -128,0 +130,4 @@ class PersistentDict(dict):
+  def __eq__(self, other):
+    if isinstance(other, PersistentDict):
+      return dict.__eq__(self, other) and self._filename == other._filename and self._transact == other._transact
+    return False
\end{diff}
\label{code:diff2}
\end{mdframed}

The above code snippet is a correct way of revising Code 1. It calls the  \mytexttt{\_\_eq\_\_} method of the super class and compares all the member variables of the subclass. Further, it does not make any unnecessary changes to the code.

\section{Design}
\label{sec:design}
Designing and tuning prompts for querying LLMs is a thriving new area of research~\cite{liu2023pre}. In this section, we describe in detail the prompt construction strategies, guided by static analysis reports. The LLM invocations in our pipeline are for generating candidate revisions, and for scoring and ranking the candidates.

\subsection{Proposer LLM: Prompting the LLM to generate code revisions}
\label{sec:proposer}
To generate code revisions for a given code quality issue and an input source file, we devise a prompt template incorporating different types of information, with elaborate natural language instructions, needed to perform the revision task. Our prompt follows the generic structure shown below, with \textit{fixed components} (\circled{p$_1$} and \circled{p$_2$}, as per configuration done in \stage{1} discussed in Section \ref{sec:overview}) as well as \emph{instance-specific components} (\circled{p$_3$}, \circled{p$_4$}, and \circled{p$_5$}) obtained dynamically:
\prompt{Proposer Prompt Template}{
\circled{p$_1$} Description of the quality issue (i.e., static check).\\
\circled{p$_2$} Rubric for resolving the quality issue. \\
\circled{p$_3$} (Optional) Relevant code blocks for doing the revision. \\
\circled{p$_4$} Input source file (in full, or localized to the block containing the issue). \\
\circled{p$_5$} Location and warning message given by the static check. \\
}

The fixed components of the prompt consist of the name of the quality check, description, and recommended ways to resolve the issue. These are provided at the time of configuring the \sysname pipeline. In our empirical evaluation in Section \ref{sec:eval}, we simply use the content from the documentation webpages~\cite{cqweb,sonarsa}. When there are missing or incomplete details about how to fix issues, we furnish the information. The instantiated prompt for our running example Code 1 is given in Figure~\ref{fig:proposer-prompt}. The text in \textit{italics} is the template, the text in {\color{teal}teal} correspond to the fixed components obtained from the tool providers, and the text in {\color{brickred}brickred} correspond to instance-specific information retrieved from static analysis (CodeQL for this example) reports. 

\begin{figure}[t]
\prompt{\genprompt (output of ``Prompt Construction'' stage in Figure \ref{fig:core})}
{
\circled{p$_1$} \textit{We are fixing code that has been flagged for the \emph{\color{teal} CodeQL} warning titled }{\color{teal}"\`{}\_\_eq\_\_\`{} not overridden when adding attributes"} \textit{which has the following description:}

{\color{teal} A class that defines attributes that are not present in its superclasses may need to override the \_\_eq\_\_() method (\_\_ne\_\_() should also be defined).

Adding additional attributes without overriding \_\_eq\_\_() means that the additional attributes will not be accounted for in equality tests.}
\newline

\circled{p$_2$} \textit{The recommended way to fix code flagged for this warning is:}

{\color{teal}Override \_\_eq\_\_ method to also test for equality of added attributes by either calling eq on the base class and checking equality of the added attributes, or implementing a new eq method that checks equality on both self and inherited attributes.}
\newline

\circled{p$_4$} \textit{Modify the Buggy code below to fix the \emph{\color{teal} CodeQL} warning(s). Output the entire code block with appropriate changes. Do not remove any section of the code unrelated to the desired fix.}
\newline

\textit{Buggy Code:}\\
{\color{brickred} \code{ class PersistentDict (dict) :\\
\ \ \ \ $\cdots$ }}
\newline

\circled{p$_5$} \textit{\emph{\color{teal} CodeQL} warning(s) for the above buggy code:}\\
{\color{brickred} The class `PersistentDict' does not override "\_\_eq\_\_" , but adds the new attributes "\_filename" and "\_transact".}
\newline

\textit{The following lines are likely to be of interest:}\\
{\color{brickred} \code{1. class PersistentDict (dict) :}}
\newline

\textit{Fixed Code:}
}
\caption{Prompt supplied to the Proposer LLM for revising Code 1. This example does not require additional relevant code blocks as context and hence, the corresponding prompt component {\protect \circled{p$_3$}} is not present.}
\label{fig:proposer-prompt}
\end{figure}

\myparagraph{Handling multiple violations in the input file:} The static analysis tool gives us the locations (lines of interest), and in some cases associated warning messages as well, where the issue was flagged in the input source file. There can be multiple locations in a single source file where the check violation is flagged. If the source file is sufficiently small (to fit in the context size of the LLM), we give the entire source file (in \circled{p$_4$}) as well as all the flagged locations and warning messages (in \circled{p$_5$}) in a single prompt. If not, we use the following procedure to instantiate the prompt(s). 

Let $\mathcal{V} = \{v_1, v_2, \dots, v_n\}$ denote the flagged locations in the input file. We start with $v_1$ and take its largest encompassing block, e.g., the class or the method containing $v_1$, that fits in the available context size. Note that this block might also encompass more violation locations $v_i$ besides $v_1$. We instantiate a prompt with this block as the input (in \circled{p$_4$}), and the subset of flagged violations (in \circled{p$_5$}). We remove this location subset from $\mathcal{V}$ and repeat. Thus, the proposer LLM is invoked with possibly multiple instantiated prompts for a single input file, and the generated code block for each prompt is patched back to the original source file appropriately.

\myparagraph{Relevant code blocks}: 
During static analysis, tools like \codeql\ identify and inspect code blocks that are relevant for determining presence/absence of a property violation. We log this information while running the static analysis and provide it as additional signal in our prompt in \circled{p$_3$}. 
For example, for a CodeQL check ``signature mismatch in overriding method'', the declaration of the overriden method from the superclass is a relevant block because the static check determines the mismatch between the overriden and overriding methods by inspecting their signatures.
On the other hand, in the example of Code 1, the CodeQL error message already provides sufficient information. As shown in Figure~\ref{fig:proposer-prompt}, \circled{p$_5$} gives the CodeQL diagnostics that \code{\_filename} and \code{\_transact} attributes are added in the subclass and are not covered by the \code{\_\_eq\_\_} method of the superclass. Thus, the fix can be constructed from local code with this diagnostic information, and there is no need for any other part of the source file. For such checks, we do not supply \circled{p$_3$}.

\subsection{Ranker LLM: Prompting the LLM to score candidate revisions}
\label{sec:verifier}
As we stated in Section~\ref{sec:overview}, static analysis tools could pass revisions that are not acceptable, e.g., introducing unintended changes or otherwise altering functional correctness of the source code. 

\prompt{\checkerprompt}{
\textit{You are an expert developer. You are verifying the code generated by LLM to fix the  warning titled }"{\color{teal} `\_\_eq\_\_` not overridden when adding attributes}" \textit{which has the following description:}
{\color{teal} A class that defines attributes that are not present in its superclasses may need to override the \_\_eq\_\_() method (\_\_ne\_\_() should also be defined). ...}
\newline

\textit{The recommended ways to fix code flagged for this warning are:}\\
{\color{teal} Override \_\_eq\_\_ method to also test for equality of added attributes by either calling eq on the base class and checking equality of the added attributes, or ...}
\newline

\textit{Your task is to assess the quality of the generated patch and rate it on the following evaluation criteria:\\
{\color{blue}Score 0}, if the patch has changes unrelated and unnecessary to fixing the warning ({\color{blue}Strong Reject}).\\
{\color{blue}Score 1}, if the patch has a few correct fixes, but still modifies the original snippet unnecessarily ({\color{blue}Weak Reject}).\\
{\color{blue}Score 2}, if the patch has mostly correct fixes but is still not ideal ({\color{blue}Weak Accept}).\\
{\color{blue}Score 3}, if the patch only makes edits that fix the warning with least impact on any unrelated segments of the original snippet ({\color{blue}Strong Accept}).}
\newline

\textit{If you find additions or deletions of code snippets that are unrelated to the desired fixes (think LLM hallucinations), it can be categorically scored 0 (Strong Reject). That said, you can make exceptions in very specific cases where you are sure that the additions or deletions do not alter the functional correctness of the code, as outlined next.}
\newline

\textit{Allowed Exceptions:\\
The following (unrelated) code changes in the diff file can be considered okay and need not come in the way of labeling an otherwise correct code change as accept (score 2 or 3). This list is not exhaustive, but you should get the idea \\
(a) deleting comments is okay, \\
(b) rewriting a = a + 1 as a += 1 is okay, even though it may not have anything to do with the warning of interest, \\
(c) making version specific changes is okay, say changing print ("hello") to print "hello".}
\newline

\textit{The following (unrelated) code changes in the diff file are NOT considered okay, and you should label the diff file as reject (score 0 or 1) even if it is otherwise correct for the query. This list is not exhaustive, but you should get the idea \\
(a) deleting or adding a print statement, \\
(b) optimizing a computation, \\
(c) changing variable names or introducing typos.}
\newline

\textit{Output only the reason and score for the patch below. Do not output anything else.}
\newline

\textit{Diff}: {\color{teal} $\langle$ diff $\rangle$}\newline
\textit{Reason}:
\newline
}

In the running example of Code 1, we see two kinds of revisions that are likely to be rejected by developers: (1) revisions that override the \eq\ method properly, but alter the functionality of the code elsewhere, such as changing the implementation of \hash, and (2) revisions that do not quite resolve the quality issue, but by-pass the \codeql\ checks anyway --- for instance, all the subclass members are explicitly enumerated in the equality check without calling \texttt{super()}.\eq for the parent members. We do not want to surface such spurious candidates to the developer. 

To this end, we use another instance of the LLM to act as a ranker that scores the candidate revisions that pass the tool, i.e., output of stage \stage{4} in Figure \ref{fig:core}, in the \sysname{} pipeline. We use a prompting strategy similar to the one used for generating the revisions themselves in the previous subsection to query the ranker LLM. The prompt template for scoring candidates is given above (some parts are elided to save space). Note that the prompt is fairly generic, and in particular, is agnostic to the type of code quality check or the static analysis tool.

In addition to information about the quality check and fix recommendation, the prompt provides description of a scoring rubric that asks the LLM to rank the revisions (from strong reject to strong accept) based on how close they are to the intended change while avoiding unrelated changes. Sometimes the LLM enforces coding conventions or styles that it encountered frequently during training and rewrites code to conform to those. We instruct the LLM to overlook such changes (indicated by ``Allowed Exceptions'' in the prompt) and provide some example scenarios. Similarly, we elaborate more on what kind of revisions should be rejected in the last part of the prompt, before providing the ``Diff'' for scoring.

\section{Experimental Setup}
\label{sec:expsetup}
\myparagraph{Datasets:} \textbf{(1)} We use a subset of the CodeQueries dataset~\cite{sahu2022learning}\footnote{\url{https://huggingface.co/datasets/thepurpleowl/codequeries}} in our experiments. It contains Python files with quality issues flagged by a set of 52 CodeQL queries (i.e., static checks). The 52 CodeQL queries are taken from the standard Python CodeQL suite; these analyze various aspects of code such as security, correctness, maintainability, and readability. In all our experiments, we use the \textbf{test split} of the CodeQueries dataset. 
Due to throttled LLM access, we are able to experiment with a subset comprising 765 files across the 52 queries.
We denote this dataset as \datasetpy. Further, we sample 10 files per query from \datasetpy\ to conduct a user study on revisions generated by \sysname. We refer to this subset as \datasetpyhuman. \textbf{(2)} We use a subset of the \sorald\  dataset~\cite{someoliayi2022sorald}\footnote{\url{https://github.com/khaes-kth/Sorald-experiments}}; this subset consists of a collection of 151 java repositories from Github (out of the 161 repositories in their full dataset) with a total of 483 files, and covers all the 10 \sq\ checks studied in \cite{someoliayi2022sorald}. We refer to this set as \datasetjava.

\myparagraph{Model configurations:} We conduct our experiments using the \textbf{\turbo{}} model as the \textbf{proposer}. We obtain $10$ responses per input source file using the OpenAI inference API. Following recent work \cite{agrawal2023guiding}, to encourage diversity in the sampled responses, we use a combination of \textit{temperature} settings for the model (that controls the stochasticity in the generated responses): 1 response with temperature = 0 (greedy decoding), 6 responses with a temperature of 0.75, and 3 responses with a temperature of 1.0. We use \textbf{\gptfour} as the \textbf{ranker} (our early investigations suggested that \gptfour is significantly better in terms of reasoning with code diffs compared to \turbo) and obtain a single response (score) per candidate revision, with temperature 0.

\myparagraph{Evaluation metrics:}
For each dataset, we report the number of files flagged and the number of total issues flagged across the files. We measure how many files have at least one revision that passes the static check and how many issues remain in files with no such revision after the Proposer LLM and Ranker LLM stages of \sysname are applied. In the user study, we measure how many files have at least one revision that is accepted by the human reviewer and report how many revisions were accepted and rejected across the files by the reviewers. We refer to the number of revisions produced by the \sysname pipeline but rejected by the human reviewer as \emph{false positives}.

\section{Evaluation}
\label{sec:eval}
Our goal is to extensively evaluate the end-to-end \sysname pipeline across various quality-improving code revision tasks and to answer the following questions:
\begin{itemize}
\item[\textbf{RQ1:}] How \textit{effective} is the \textit{end-to-end }\sysname pipeline in mitigating code quality issues and in passing scrutiny by the Ranker LLM on the Python benchmark \datasetpy? 
\item[\textbf{RQ2:}] How many of the \sysname-generated revisions are also \emph{accepted by human reviewers} on the Python benchmark \datasetpyhuman?
\item[\textbf{RQ3:}] How \textit{readily} does \sysname pipeline \textit{generalize} to a different programming language (Java) and a static analysis tool (SonarQube)?
\item[\textbf{RQ4:}] How \textit{well} does \sysname \textit{compare} to a state-of-the-art automatic program repair technique (\sorald) for mitigating static analysis warnings?
\end{itemize}

\subsection{RQ1: How \textit{effective} is the \textit{end-to-end }\sysname pipeline in mitigating code quality issues and in passing scrutiny by the Ranker LLM on the Python benchmark \datasetpy?}
\label{sec:RQ1}
We start by looking at the overall performance of the \sysname\ pipeline in terms of (1) fixing the code quality issues as determined by the static analysis tool that \sysname is configured with, and (2) acceptances as determined by the Ranker LLM using a detailed evaluation criteria to assess the code revisions (as described in Section \ref{sec:verifier}).  

The overall evaluation results of the end-to-end \sysname\ pipeline (on the datasets and metrics introduced in Section \ref{sec:expsetup}) are presented in Table \ref{tab:codequeriesrq1}. For RQ1, we will focus on the first row, that corresponds to \sysname pipeline configured with \codeql\ as the static analysis tool, and the 52 quality checks that are part of the \datasetpy\ Python dataset. 

The first block of columns shows the dataset statistics. There are 1993 quality issues (i.e., static check violations) flagged in the 765 files of the \datasetpy\ dataset, with each file having at least one issue flagged, by the end of stage \stage{1} in Figure \ref{fig:core}. In the second block of columns, we show the effectiveness of the Proposer LLM, \textit{after} the proposed candidate revisions (10 revisions per flagged file) are filtered by the tool (i.e., \codeql\ for the first row) by the end of stage \stage{4}. First, we observe that 81.57\% of the flagged files get fixed entirely as validated by the static analysis tool, i.e., they have at least one revision that completely passes the static checker with zero issues flagged. Second, we observe that, the average number of issues remaining per revised file, by the end of stage \stage{4} of the \sysname pipeline, is 0.41 compared to over 2.6 issues on average per source file at the beginning of the pipeline. This is particularly remarkable as the Proposer LLM is able to perform revision with just the natural language instructions, without explicitly providing any training examples of the form $\langle$\textit{before} code, \textit{after} code$\rangle$ that are commonly needed for automatic program repair tools. 

The instruction-following ability of the Proposer LLM to do code revisions, although impressive, can also produce spurious fixes that pass static checks. In the last stage of the \sysname pipeline, the Ranker LLM uses elaborate evaluation criteria (in its carefully-constructed prompt presented in Section \ref{sec:verifier}) to reject such spurious fixes and accept revisions that are likely to be also accepted by developers. From the last block of columns of Table~\ref{tab:codequeriesrq1}, we see that 583 out of 624 files are ranked high, i.e., strong or weak accept, by the Ranker LLM by the end of stage \stage{5}. In particular, the Ranker LLM (strong- or weak-) rejects every revision (possibly spurious) for 41 files even though they are passed by the tool in stage \stage{4}. In the subsequent RQ, we analyse how well the acceptances and the rejections by the Ranker LLM correlate with human reviewers, on a subset of the \datasetpy\ dataset.

\begin{table*}[t] 
\centering 
\begin{tabular}{@{}l@{\;}|@{\;}r@{}r@{\;}|@{}r@{}r@{\;}|r@{}r@{}}
\toprule 
\small Dataset & \multicolumn{2}{c|}{\small Dataset statistics} & \multicolumn{2}{c|}{\small Effectiveness of Proposer LLM} & \multicolumn{2}{c}{\small Rankings by Ranker LLM}\\  
\cmidrule(l{0em}r{0em}){2-7} 
& \footnotesize {\#Files flagged} & \multicolumn{1}{c|}{\footnotesize{\#Issues flagged}} & \multicolumn{1}{c}{\footnotesize \#Files passing} & \multicolumn{1}{c|}{\footnotesize \#Issues remaining} & \multicolumn{2}{c}{\footnotesize \#Files ranked}\\  
& &  \multicolumn{1}{c|}{\footnotesize (Avg. per file)} & \multicolumn{1}{c}{\footnotesize static checks (\%)} & \multicolumn{1}{c|}{\footnotesize (Avg. per file)} & \footnotesize high (\%)& \footnotesize low (\%)\\ 
\toprule
\small\datasetpy & \small 765 \footnotesize(100\%) & \small 1993 \footnotesize(2.61)  & \small 624 \footnotesize{(81.57\%)} & \small 315 \footnotesize{(0.41)} & \small 583 \footnotesize (76.21\%) & \small 41 \footnotesize(5.36\%)  \\
\small\datasetpyhuman & \small 520 \footnotesize(100\%) & \small 999 \footnotesize{(1.90)} & \small  453  \footnotesize{(87.11\%)} & \small 159 \footnotesize{(0.31)} & \small 427 \footnotesize{(82.11\%)} & \small 26 \footnotesize{(5.00\%)}\\ 
\small\datasetjava & \small 483 \footnotesize(100\%) & \small 999 \footnotesize{(2.06)} & \small  397 \footnotesize{(82.19\%)} & \small 270 \footnotesize{(0.56)} & \small 371 \footnotesize{(76.81\%)}  & \small 26 \footnotesize{(5.38\%)}\\
\bottomrule
\end{tabular} 
\caption{Summary of end-to-end evaluation of \sysname\ on real-world Python and Java files, with 52 and 10 static checks using \codeql\ and \sq\ respectively. ``\#Files flagged'' and ``\#Issues flagged'' correspond to output of static checks (stage {\protect \stage{1}} in Figure~\ref{fig:core}). ``\#Files passing static checks'' and ``\#Issues remaining'' report the number of files having at least one revision that passes the static checks and the issues that remain in files with no such revision (stage {\protect \stage{4}} output). ``\#Files ranked high (low)'' is the number of files with at least one revision (no revision, respectively) that is scored as weak/strong accept by the Ranker LLM (stage {\protect \stage{5}} output). For files, the percentages are reported with respect to the ``\#Files flagged''.} 
\label{tab:codequeriesrq1} 
\end{table*}

\subsection{RQ2: How many of the \sysname-generated revisions are also \emph{accepted by human reviewers} on the Python benchmark \datasetpyhuman?}
\label{sec:RQ2}

\begin{table}
\centering
\begin{tabular}{@{}l|rr|rrr@{}}
\toprule
Stage evaluated & \multicolumn{2}{c|}{Stage-wise output} & \multicolumn{3}{c}{Results of user study}\\
\cmidrule(l{0em}r{0em}){2-6}
& \multicolumn{1}{c}{\footnotesize \#Files} & \multicolumn{1}{c|}{\footnotesize \#Revisions} & \multicolumn{1}{c}{\footnotesize \% Files} & \multicolumn{1}{c}{\footnotesize \% Revisions} & \multicolumn{1}{c}{\footnotesize \% Revisions} \\
& \multicolumn{1}{c}{\footnotesize retained} & \multicolumn{1}{c|}{\footnotesize retained} & \multicolumn{1}{c}{\footnotesize accepted (\#)} & \multicolumn{1}{c}{\footnotesize accepted (\#)} & \multicolumn{1}{c}{\footnotesize rejected (\#)}\\
\midrule
\small Stage \stage{4} (Proposer LLM) & \small 453 \footnotesize{(100\%)}  & \small 2397 \footnotesize(100\%) &  \small 70.64\% \footnotesize{(320)} & \small 44.89\% \footnotesize{(1076)} &  \small 55.11\% \footnotesize{(1321)}\\ \hline 
\small Stage \stage{5} (Ranker LLM, {\color{teal} SA}) & \small 410 \footnotesize(100\%) & \small 1756 \footnotesize(100\%) & \small \textbf{\small 72.68\%} \footnotesize{(298)} & \textbf{\small 52.45\%} \footnotesize{(921)} & \small 47.55\% \footnotesize{(835)}\\
\small Stage \stage{5} (Ranker LLM, {\color{teal} WA}) & \small 17 \footnotesize(100\%) & \small 228 \footnotesize(100\%) & \small 58.82\% \footnotesize{(10)} & \small 36.40\% \footnotesize{(83)} & \small 63.60\% \footnotesize{(145)}\\
\small Stage \stage{5} (Ranker LLM, {\color{red} WR/SR}) & \small 26 \footnotesize(100\%) & \small 413 \footnotesize(100\%) & \small 46.15\% \footnotesize{(12)} & \small 17.43\% \footnotesize{(72)} & \small \textbf{\small 82.57\%} \footnotesize{(341)}\\
\bottomrule
\end{tabular}
\caption{Results of user study on the \datasetpyhuman\ dataset. ``Ranker LLM, {\color{teal}SA}'' denotes all the revisions scored as strong accept by the Ranker LLM; ``Ranker LLM, {\color{teal} WA}'' denotes all the revisions scored as weak accept by the Ranker LLM, and ``Ranker LLM, {\color{red} WR/SR}'' denotes all the revisions scored as rejects (strong or weak) by the Ranker LLM. For files and revisions, the percentages are reported row-wise with respect to the numbers in the first block of columns (under ``Stage-wise output''). Column-wise maximums are in the bold typeface.}
\label{tab:user-study}
\end{table}

In this RQ, we investigate the correctness of the revisions produced by \sysname, and in particular the effectiveness of the Ranker LLM, by conducting a user study. We use a subset of \datasetpy, called \datasetpyhuman, with a sample of 10 files per quality check, which already yields 2397 candidate revisions (out of stage \stage{4}) to be manually scrutinized. The \sysname pipeline results for this dataset are presented in the second row of Table \ref{tab:codequeriesrq1}, where the trend closely resembles that of \datasetpy\ in the first row. 

For each of the 453 files in \datasetpyhuman\ that comes out of stage \stage{4} (as seen from row 2, Table \ref{tab:codequeriesrq1}), we ask a human reviewer to label all the revisions for the file as \textit{accept} or \textit{reject}. We provide the same rubric that we give as prompt to the Ranker LLM (presented in Section \ref{sec:verifier}) to the reviewer to assess the correctness of the revisions --- the only change is that we ask the reviewer to give a binary accept/reject decision than a graded score that we elicit from the Ranker LLM. Our user group consists of 15 Python developers (intermediate level, with 1-3 years of software engineering experience). Each revision was labeled by only one user, and each user was responsible for labeling revisions of 2 to 4 (randomly chosen) queries from the dataset. None of the authors of this paper were part of the user group.

The results of the \datasetpyhuman\ user study are presented in Table \ref{tab:user-study}. The first row of the Table shows the (baseline) metrics for the user study we conducted --- all the outputs of stage \stage{4} were reviewed by the users, and 70.64\% of the reviewed files have at least one revision that a human reviewer accepted. However, from the last column, we see that this high acceptance rate comes at a high cost of 1321 false positives, i.e., 55.11\% of the revisions that \codeql\ passed were rejected by users. This trade-off between acceptance rate and false positives of the pipeline can be crucial in practice. In the following, we show that the Ranker LLM helps achieve a significantly better trade-off. 

Equipped with the accept/reject labels given by users for all the \codeql-passed revisions of the \datasetpyhuman\ dataset, we ask: \textit{Can the Ranker LLM help tell the correct revisions from the incorrect ones, which would in turn help minimize the review burden of developers?} We answer this question affirmatively in the subsequent rows of Table \ref{tab:user-study}. From the second row, we see that if we surface only the candidates strongly accepted by the Ranker LLM, the rejection rate drops to 47.55\%. In an absolute sense, the number of rejections drops to 835 from 1321, which is close to 25\% reduction. At the same time, 72.68\% of the scrutinized files have at least one revision accepted by a reviewer. Furthermore, from the last row, we see that if we consider only the files for which no revision was (strong- or weak-) accepted by Ranker LLM, the users also rejected over 82\% of those revisions; this indicates that dropping the low confidence rejections by the Ranker LLM can indeed help significantly reduce the review burden of developers in practice.

\begin{framed}\noindent\sysname\ \textit{pipeline, with the Proposer-Ranker duo LLMs, for resolving code quality issues is effective and can be deployed in real software engineering workflows. Relying only on the (symbolic) tools for filtering revisions is problematic --- the Ranker LLM helps reduce the number of false positives greatly, while also ensuring that acceptable revisions which preserve functional correctness are surfaced to the developers.}\end{framed}

\subsection{RQ3: How \textit{readily} does \sysname pipeline \textit{generalize} to a different programming language (Java) and a static analysis tool (SonarQube)?}
\label{sec:RQ3}
\sysname can handle different programming languages and static analysis tools out of the box. To demonstrate this, in this RQ, we configure \sysname\ with another widely-used static analysis tool \sq, and the 10 static checks from the \datasetjava\ dataset (introduced in Section \ref{sec:expsetup}). This configuration was straight-forward; \textit{it took us less than a week to get this done}. In fact, lines of code that needed changes in our \sysname implementation (in Python) for this configuration was less than 100. Specifically, we did \textit{not} have to adapt or tune the prompts of the Proposer and Ranker LLMs in our pipeline to accommodate the new tool or the programming language. The authors of the \sorald\ dataset have made available clear descriptions and fix recommendations for the 10 checks, which we readily use to instantiate our LLM prompts. Further, \sq\ provides localization for the check violations (line numbers in the source file) needed to extract code blocks as discussed in Section \ref{sec:proposer}.

We report results on the \datasetjava\ dataset consisting of real-world Java repositories in the last row of Table \ref{tab:codequeriesrq1}. There are 999 quality issues flagged in 483 files of the dataset, with each file having at least one issue flagged, by the end of stage \stage{1} of \sysname. As in the case of the other datasets (first and second rows), we find that over 82\% files have at least one candidate revision that entirely passes the associated \sq\ check. Furthermore, the average number of issues that remain by the end of stage \stage{4} is about 0.56 per file, compared to over 2 issues per file on average to begin with. From the last column, we see that the Ranker LLM rejects all the (possibly spurious) revisions that passed \sq\ checks for 26 files, and (strong- or weak-) accepts at least one revision for 371 files, which is over 76\% of the total files.

\subsection{RQ4: How \textit{well} does \sysname \textit{compare} to a state-of-the-art automatic program repair technique (Solard) for mitigating static analysis warnings?}
\label{sec:RQ4}

\begin{table*}[h] 
\centering 
\begin{tabular}{crr} \toprule 
 & \#Files (\%) & \#Issues remaining (\%) \\ \midrule
 Flagged & 483 (100\%) & 999 (100\%)\\
 \sysname  & 371 (76.8\%) & 270  (27.03\%)  \\
 \sorald & 378 (78.3\%) & 371 (37.14\%)  \\ 
\bottomrule
\end{tabular} 
\caption{Comparison of \sysname with the state-of-the-art automatic program repair method \sorald on the \datasetjava\ dataset consisting of 10 static checks, using \sq\ as the static analysis tool. ``\#Files (\%)'' for \sysname and Solard rows indicate the number of files (\% with respect to the Flagged files) that are fixed by the tools respectively.} 
\label{tab:sonarquberq4} 
\end{table*}
We compare \sysname with the state-of-the-art automatic program repair tool \sorald \cite{someoliayi2022sorald} for fixing static check issues in code. \sorald is a rule-based approach that leverages ``metaprogramming templates'', which are basically AST-to-AST transformations, that can be applied on the detected violations in code. In particular, for each violation location in the code, \sorald applies one metaprogramming template to the corresponding AST element to fix it. They manually implement one metaprogramming template per static check, based on the fix recommendations for the check, which we directly use in the form of natural language instructions in the \sysname pipeline. While their repair tool is extensible to other languages and static analysis tools, their publicly available implementation\footnote{\url{https://github.com/ASSERT-KTH/sorald}} is for Java and \sq. So, for this RQ, we focus on the \datasetjava\ dataset.

The comparison results on the Java dataset are presented in Table \ref{tab:sonarquberq4}. Of the 483 files in the dataset, \sysname, i.e., the output of stage \stage{5}, considering the (strong/weak) accepted revisions by the Ranker LLM, fixes 371 files entirely, at a rate of 76.8\%. This is comparable to the manually crafted \sorald tool that fixes 378 files. On the other hand, the number of issues that remain by the end of \sysname\ pipeline is 270 (about 27\%), significantly less compared to 371 (over 37\%) for the \sorald\ tool. 

\section{Threats to validity}
A possible threat to validity is that the input code in our dataset might have been seen by the LLM during its training. 
The LLM is unlikely to have seen the prompts constructed by us paired with the expected code revisions during training. Therefore, our results can be attributed to the ability of the LLMs to follow the instructions, their knowledge of programming languages and the informative details we provide in our prompts. By basing our experiments on hundreds of issues flagged by 52 diverse static checks for Python and 10 diverse static checks for Java from two different static analysis tools, we avoid the possibility of biasing our results to a small dataset, certain code quality issues, or a single tool or programming language.
We follow the exact experimental setup as CodeQueries and Solard to avoid any language or tool version mismatch issues.

The code generated by the LLM may pass the previously failing static checks but change the code semantics, e.g., by completely deleting the code. 
To mitigate this problem, we perform human evaluation for verifying soundness of the revisions, albeit on a subset of our Python dataset, but ensuring full coverage in terms of the static checks. This manual labeling could be noisy. All the labels were independently reviewed by one of the authors to avoid such cases. 

We found cases where the users were unsure why the tool flagged a violation in the source code in the first place, or whether the fix in the revision had no unintended side effects.  There were also a few cases that proved to be challenging to manually verify the correctness of the revisions. For instance, consider the ``\code{import *} may pollute namespace'' static check for Python files\footnote{https://codeql.github.com/codeql-query-help/python/py-polluting-import/}. The correct revisions would replace the \code{*} with relevant modules. However, verifying if all the required imports are fully enumerated can be challenging, especially for large source files. Looking at multiple revision candidates for some files was helpful to users in this regard --- whenever two candidates for a file had a non-overlapping subset of enumerated imports, the user tried to reason about the differences and was able to resolve incompleteness of one or both the revisions. To avoid cases from inflating or otherwise biasing our evaluation results, we instructed the users to \textit{reject} revisions that they were unsure of, as in some of the examples mentioned above, erring on the safer side.

\section{Related work}
Automatic program repair is a topic of active research and many tools have been built over the years. Here, we discuss the most closely related work and refer the reader to excellent surveys~\cite{monperrus2018automatic,GouesPradelRoychoudhury2019,huang2023survey} for wider coverage.

\subsection{Repairing static check violations} Among the approaches that target static analysis errors, \cite{FootPatch,senx,someoliayi2022sorald,costea2023hippodrome} use manually designed symbolic program transformations to fix specific classes of properties like heap safety~\cite{FootPatch}, security vulnerabilities~\cite{senx}, static quality checks~\cite{someoliayi2022sorald} or data races~\cite{costea2023hippodrome}. Other approaches~\cite{sonarcube,Liu:mining,8667970,rolim2018learning,bader2019getafix,bavishi2019phoenix} mine symbolic patterns from commit data to learn repair strategies or learn them from synthetically generated data~\cite{jain2023staticfixer}.
For instance, \othersysname{SpongeBugs}~\cite{sonarcube} uses SonarQube~\cite{sonarsa} to find bugs and commit data to create paired dataset.
Similarly,~\othersysname{Avatar}~\cite{Liu:mining,8667970} and \othersysname{Phoenix}~\cite{bavishi2019phoenix} use FindBugs~\cite{findbugs} and commit data. \othersysname{Revisar}~\cite{rolim2018learning} mines edits from commit data for PMD~\cite{pmd}.
\othersysname{GetAFix}~\cite{bader2019getafix} uses Infer~\cite{isa} and Error Prone~\cite{aftandilian2012building}, and mines general tree edit patterns from commit data using anti unification. 

These repair techniques can synthesize only those fixes that are covered by their symbolic patterns. An alternative approach based on learning~\cite{tufano2019empirical,zhu2021syntax,jiang2021cure,ye2022neural} is to train neural models to map buggy programs to their fixed versions. The neural models learn to directly transform code. However, their scope is determined by the diversity of bug-fixing examples present in the training data and they do not generalize to new classes of bugs not seen during training.

All these approaches require extensive data curation and offline learning efforts, and require redesign when targeting different kinds of bugs. In contrast, the line of work we pursue, using LLMs, does not require any data curation or learning effort. Since LLMs have aleady been pretrained with a large corpus of code and other documents, they can be readily customized to revise code to fix any type of error detected by static analysis, just by suitably authoring prompts.

\subsection{LLMs for program repair} The aforementioned advantage of using LLMs has motivated other researchers to use them for program repair. Xia, Wei and Zhang~\cite{xia2023automated} use LLMs with few-shot prompts to generate candidate fixes on buggy code from Defects-4J, QuixBug and ManyBugs benchmarks and use entropy values (the negative log probability of each generated token) to rank candidate fixes. The work relies on the existence of a test suite to validate a candidate fix.
In a more recent work, Xia and Zhang~\cite{xia2023conversational} use a conversational approach, where a test suite is a requirement, and error messages from failed tests are used in a conversational style with the LLM to refine the candidate fix into one that passes the test suite, and present results on the QuixBug benchmarks.
Another interesting line of work is to fix bugs in code generated by an LLM using traditional program repair techniques or another LLM~\cite{jain2022jigsaw,fan2022automated,liventsev2023fully} .

These approaches aim at fixing bugs identified by failing test cases. In comparison, our work addresses a related but different problem, one of fixing errors flagged by static analysis tools such as CodeQL and SonarQube. 

\myparagraph{Prompting techniques:} RING~\cite{joshi2022repair} fixes syntactic and simple semantic errors across multiple languages using an LLM and retrieval-augmented few-shot prompting. The complexity of errors and required fixes in our case is higher. InferFix~\cite{jin2023inferfix} targets violations flagged by the Infer static analyzer~\cite{isa,isharp} for three types of bugs. However, it constructs prompts augmented with bug type annotation and similar bug-fix pairs, and finetunes the Codex model on these prompts. We use an instruction-based LLM in zero-shot setting (i.e., no $\langle$\textit{before} code, \textit{after} code$\rangle$ examples needed) without finetuning. Pearce et al.~\cite{pearce2022examining} fix security vulnerabilities using auto-regressive LLMs which are prompted with partial code in which the buggy lines are commented out and the LLM is prompted to generate a ``fixed" version of those. We use a more powerful class of instruction-tuned LLMs which benefits from detailed instructions that provide additional context necessary for generating correctly revised code. Our prompts encompass description of the quality issue, suggested resolutions, localization hints and constraints. Due to the auto-regressive nature, the generations in~\cite{pearce2022examining} are conditioned only on prefix of the buggy code, whereas we pass the buggy code in the prompt and hence, the code generation can attend to the bidirectional code context, both before and after the buggy lines in the input code.

\myparagraph{Validating fixes:} The above approaches rely on static analysis to validate the fixes and availability of unit tests to detect regressions. We use a combination of two oracles: (1) the static analyzer, and (2) a second ranker LLM, to ensure that the fix does not vacuously pass the static analysis, to automatically generate acceptable fixes that are both functionally correct and pass static analysis. 

\subsection{LLMs as verifiers}
The issue of plausible but incorrect fixes is well-known~\cite{perkins2009automatically,long2015staged}. CORE may generate code that passes the static check (a plausible fix) but changes semantics of the input code in unintended ways. The developer can review the statically-validated code-revisions to filter out such cases. Unit tests can also help catch such cases, but they are not always available or may themselves be incomplete. LLMs have been shown to be effective in assessing and supervising quality of output from other LLMs~\cite{bai2022constitutional,kocmi2023large}, thereby helping reduce the efforts required for human review. Using LLMs, especially GPT-4, for evaluating code generations has been attempted recently. Olausson et al. \cite{olausson2023demystifying} also have a dual LLM setup, where they use the feedback from GPT-4 in the form of critique  to modify the prompt of the proposer LLM for code generation tasks. Zhuo \cite{zhuo2023large} constructs an elaborate prompt for \turbo\ to perform two aspects of evaluation of code generations, namely, code usefulness and evaluation-based functional correctness. Inspired by these findings, to reduce the burden on the developer, we employ a second instance of LLM (\gptfour) as a ranker to score the candidates produced by the proposer LLM based on (1) the correctness of issue resolution, and (2) preserving the functional correctness of the original code. The code generation datasets studied in \cite{zhuo2023large} consist mostly of small code snippets, unlike our setting where we use large real source code files. We work with code diffs in the Ranker LLM prompt, and in our investigations, \gptfour is substantially better in terms of reasoning with code diffs compared to \turbo that Zhuo \cite{zhuo2023large} employs.

\section{Conclusions and Future Work}
Code quality is a persistent concern in software engineering. Though much progress has been made in detecting these issues statically, fixing them automatically has remained challenging due to the variety of code quality issues that surface in real code. Our proposal in this work is to use the power of large language models, particularly, those that go beyond code completion and can follow natural language instructions, to assist developers in revising and improving their code. Through comprehensive evaluation on two public benchmarks in Python and Java that use 52 and 10 static checks from two different tools, we show the promise of this approach when coupled with carefully crafted prompts. We further show that by employing an LLM instance as a ranker, that assesses the likelihood of acceptance of proposed code revisions, we can effectively catch plausible but incorrect fixes and reduce developer burden.

Our objective for future is to expand the scope of our tool \sysname by building more components in the pipeline to not only support more tools and checks but to also improve the quality and correctness of the generated fixes. We believe that feedback-driven continuous improvement is a key to make this work mainstream. For this, we plan to draw upon the traditional static and dynamic analysis techniques for automated feedback generation and use the recent advances in finetuning of the models using techniques based on reinforcement learning and human feedback~\cite{christiano2017deep,stiennon2020learning,ouyang2022training}. 

\bibliographystyle{ACM-Reference-Format}
\bibliography{references}

\end{document}